\documentclass[runningheads]{llncs}

\usepackage[year=2026]{eccv}
\usepackage{eccvabbrv}

\usepackage{graphicx}
\usepackage{booktabs}
\usepackage{pgfplotstable}
\usepackage{siunitx}
\usepackage[accsupp]{axessibility}
\usepackage{hyperref}
\usepackage{orcidlink}
\usepackage{wrapfig}

\usepackage{tikz}
\usepackage{tkz-euclide}
\usetikzlibrary{calc}
\usetikzlibrary{math}
\usetikzlibrary{shapes}
\usetikzlibrary{decorations.pathreplacing}

\newcommand{\MITMedianLLGain}{0.21}
\newcommand{\MITMaxLLGain}{0.73}
\newcommand{\MITMedianAUCGain}{1.1}
\newcommand{\MITMaxAUCGain}{5.3}
\newcommand{\COCOMedianLLGain}{0.26}

\newcommand{\COCOMedianAUCGain}{1.7}

\newcommand{\VideoOldIOC}{2.00}
\newcommand{\VideoNewIOC}{3.05}

\newcommand{\VideoUniformerSalLL}{1.91}

\newcommand{\LFNStimuli}{75}
\newcommand{\LFLOSOCrossval}{2.22}
\newcommand{\LFLOFOCrossval}{2.25}
\newcommand{\LFGapLOSOLOFO}{0.04}
\newcommand{\LFGapLOSOLOFONoDG}{0.06}
\newcommand{\LFDGLeakage}{0.014}

\newcommand{\LFUpperOverfitLOSO}{0.30}
\newcommand{\LFUpperOverfitLOFO}{0.35}
\newcommand{\LFUpperOverfitNoDGLOSO}{0.52}

\begin{document}
\pagestyle{plain}

\title{Raising the Ceiling: Better Empirical Fixation Densities for Saliency Benchmarking}

\titlerunning{Raising the Ceiling: Better Empirical Fixation Densities}

\author{Susmit Agrawal\inst{1} \and
Jannis Hollmann\inst{1} \and
Matthias K\"ummerer\inst{1}}

\authorrunning{S.~Agrawal et al.}

\institute{Tübingen AI Center, University of T\"ubingen, Germany}

\maketitle

\begin{abstract}
Empirical fixation densities, spatial distributions estimated from human eye-tracking data, are foundational to saliency benchmarking.
They directly shape benchmark conclusions, leaderboard rankings, failure case analyses, and scientific claims about human visual behavior. Yet the standard estimation method, fixed-bandwidth isotropic Gaussian KDE, has gone essentially unchanged for decades. This matters now more than ever: as the field shifts toward sample-level evaluation (failure case analysis, inverse benchmarking, per-image model comparison), reliable per-image density estimates become critical.

We propose a principled mixture model that combines an adaptive-bandwidth KDE based on Abramson's method, center bias and uniform components, and a state-of-the-art saliency model,
to capture different spatial and semantic types of interobserver consistency, and optimize all parameters per image via leave-one-subject-out cross-validation. Our method yields substantially higher interobserver consistency estimates across multiple benchmarks, with median per-image gains of 5–15\% in log-likelihood and up to 2 percentage points in AUC. For the most affected images --- precisely those most relevant to failure case analysis --- improvements exceed 25\%. We leverage these improved estimates to identify and analyze remaining failure cases of state-of-the-art saliency models, demonstrating that significant headroom for model improvement remains. More broadly, our findings highlight that empirical fixation densities should not be treated as fixed ground truths but as evolving estimates that improve with better methodology.
\end{abstract}

\section{Introduction}
\label{sec:intro}

Humans sample visual information through sequences of fixations. While scanpath dynamics are informative~\cite{kleinInhibitionReturn2000,boccignoneModellingGazeShift2004,tatlerSystematicTendenciesScene2008,smithLookingBackWaldo2011,engbertSpatialStatisticsAttentional2015,tatlerLATESTModelSaccadic2017,hoppeMultistepPlanningEye2019,kadnerImprovingSaliencyModels2023}, for many purposes it suffices to ignore the sequential structure and treat fixations as samples from an unknown probability distribution over locations \cite{ittiComparisonFeatureCombination1999,hendersonHumanGazeControl2003,baddeleyHighFrequencyEdges2006,vincentWeLookLights2009,tsengQuantifyingCenterBias2009,kummererPredictingVisualFixations2023}. This \emph{empirical fixation density}---our best estimate of the spatial structure shared across observers---underpins saliency benchmarking, model evaluation, and scientific claims about visual attention.

\begin{wrapfigure}{r}{0.52\textwidth}
    \centering
    \vspace{-0.0cm}
    \begin{tikzpicture}
        \coordinate (hsep) at (-0.2, 0);
        \coordinate (vsep) at (0, -0.1);
        \coordinate (labelsep) at (0.2, -0.5);
        \tikzset{label/.style={font=\sffamily\bfseries}}
        \tikzset{anchor=north west}
        \node (case) at (0, 0) {\includegraphics[]{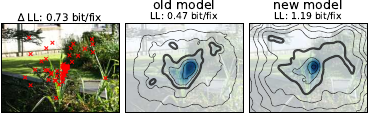}};
        \node (benchmark) at ($ (case.south west) + (vsep) $) {\includegraphics[]{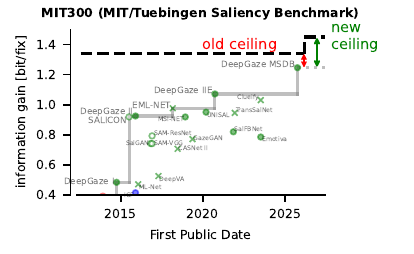}};
        \node[label, anchor=south] at ($ (case.north west) + (labelsep) $) {(a)};
        \node[label, anchor=south] at ($ (benchmark.north west) + (labelsep) $) {(b)};
    \end{tikzpicture}
    \vspace{-0.9cm}
    \caption{(a) We propose a new model of empirical fixation densities that shows that existing methods sometimes substantially underestimated interobserver consistency. (b) This restores headroom on saliency benchmarks for future progress (see Appendix~\ref{app:benchmark_progress} for details).
    }
    \vspace{-0.7cm}
    \label{fig:teaser}
\end{wrapfigure}
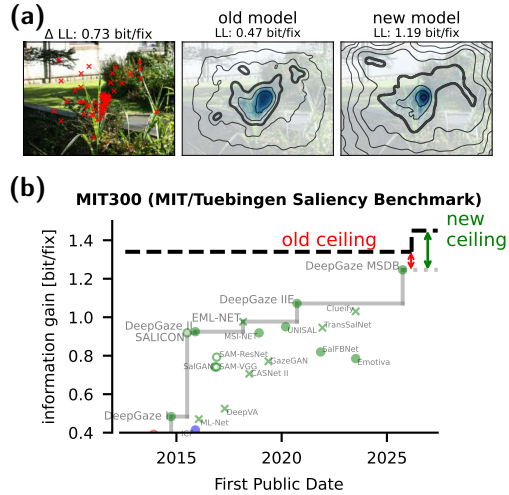
Empirical fixation densities are not only used to report average benchmark performance; they increasingly drive \emph{how} we reason about models. First, they define the \emph{interobserver consistency} (IOC): the leave-one-subject-out predictive performance of human fixations against the density estimated from other observers \cite{wilmingMeasuresLimitsModels2011,dorrVariabilityEyeMovements2010}, which acts as the ceiling against which computational saliency models are compared. Second, they enable case studies and failure analysis by supporting reliable \emph{per-image} comparisons---for example, identifying images where models systematically disagree with humans and characterizing the failure modes behind these errors \cite{kummererUnderstandingLowHighLevel2017,bylinskiiWhereShouldSaliency2016,linardosDeepGazeIIECalibrated2021,kummererModelingSaliencyDataset2025}. Third, they are frequently used as stand-ins for saliency maps in downstream models of gaze and attention, including scanpath models, to avoid confounders introduced by imperfect intermediate saliency predictors \cite{engbertSpatialStatisticsAttentional2015,schwetlickModelingEffectsPerisaccadic2020}.

Crucially, the quality of an empirical density estimate can affect more than a single topline number. For metrics computed directly from fixation locations (e.g., log-likelihood, AUC, NSS), the estimated density primarily determines only the \emph{ceiling estimate} (IOC), while model scores remain tied to the raw fixation data. In contrast, for also commonly used map-comparison metrics (e.g., CC, KL-divergence), the empirical density is itself the target map; changing the density changes model scores and can therefore affect rankings and scientific conclusions about which models are better \cite{kummererSaliencyBenchmarkingMade2018}.

This dependence becomes more consequential as the field shifts from aggregate leaderboard comparisons toward \emph{sample-level} evaluation \cite{ghoshONEBenchTestThem2025}: per-image analyses \cite{volokitinPredictingWhenSaliency2016}, inverse benchmarking~\cite{dagostinoWhatMovesEyes2025} (finding stimuli of maximal model error), and fine-grained failure mode characterization \cite{bylinskiiWhereShouldSaliency2016}. These settings require reliable per-image gold standards. If density estimates vary substantially in quality across images, then sample-level conclusions can become unstable precisely for the images where detailed analysis matters most.

Beyond benchmarking, interobserver consistency plays a broader methodological role: it provides an estimate of the \emph{explainable variance} in gaze behavior \cite{kriegeskorteRepresentationalSimilarityAnalysis2008,kummererInformationtheoreticModelComparison2015}. Statistical significance can establish that an effect exists, but comparing an effect's explanatory power to the explainable variance indicates whether it is behaviorally meaningful. In practice, it is possible to obtain highly significant results that explain only a small fraction of what is explainable; without a reliable topline, such results are difficult to calibrate.
This calibration also benefits from sample-level analysis. Effects that are small on average may be highly consequential for specific subsets of images (e.g., rare but semantically important events). Both the topline estimate and its sample-level decomposition depend on accurate empirical fixation densities.

Despite the importance of accurate density estimates across all of these settings, the estimation procedure has changed remarkably little. Bandwidth choice is typically justified heuristically---often motivated by foveal size~\cite{tatlerCentralFixationBias2007} and ``commonly accepted'' to be one degree of visual angle~\cite{meurMethodsComparingScanpaths2013}, despite there being ``no governing rule''~\cite{caldaraIMapNovelMethod2011}.
This status quo has become a practical bottleneck. State-of-the-art saliency models now match or exceed classical gold standard estimates on several benchmarks~\cite{kummererModelingSaliencyDataset2025}. Since interobserver consistency is intended as a ceiling on achievable prediction performance, a model surpassing this ceiling signals that the ceiling estimate itself is too low: the density estimation method, not human consistency, is the limiting factor. When the ceiling is too low, both benchmark interpretation and sample-level failure analysis become unreliable.

We introduce a principled mixture model framework for estimating empirical fixation densities that combines adaptive-bandwidth KDE based on Abramson's method with center-bias and uniform components as spatial priors and a state-of-the-art saliency model as a semantic regularizer, with all parameters optimized per image via leave-one-subject-out (LOSO) cross-validation. Our contributions are: (i)~the first application of adaptive-bandwidth KDE (Abramson's method) to fixation density estimation, capturing multi-scale spatial structure; (ii)~a mixture-model framework combining spatial and semantic components with per-image optimization; (iii)~comprehensive empirical analysis showing substantially higher interobserver consistency across multiple datasets and the MIT/T\"ubingen Saliency Benchmark holdout sets; and (iv)~per-image analysis revealing even larger improvements in the tail, which is crucial for sample-level benchmarking and case studies. We additionally introduce an improved density visualization method for meaningful qualitative comparison of fixation distributions. Our results demonstrate that interobserver consistency is not a fixed property of a dataset but depends critically on the quality of density estimation.%

\section{Related Work}
\label{sec:related}

\subsubsection{Fixation Density Estimation and Saliency Evaluation}
\label{sec:related:density}

Kernel density estimation has been the standard method for constructing empirical fixation densities since at least \cite{woodingEyeMovementsLarge2002}.%
The most common choice is a fixed-bandwidth isotropic Gaussian kernel, with bandwidth typically set to one degree of visual angle~\cite{meurMethodsComparingScanpaths2013,gideLocallyWeightedFixation2016,bylinskiiWhatDifferentEvaluation2018} or derived from rules of thumb such as Scott's rule~\cite{engbertSpatialStatisticsAttentional2015,rothkegelTemporalEvolutionCentral2017}. Caldara and Miellet~\cite{caldaraIMapNovelMethod2011} note the absence of a principled bandwidth selection criterion.

Due to historical reasons~\cite{ittiModelSaliencybasedVisual1998}, fixation prediction models traditionally output \emph{saliency maps}, which were originally conceived as internal representations rather than quantitative predictions and whose relationship to actual fixation behavior was only loosely specified. Different evaluation metrics (e.g.~AUC, NSS, CC, KL-divergence) imposed different implicit assumptions and often yielded contradictory rankings~\cite{richeSaliencyHumanFixations2013,wilmingMeasuresLimitsModels2011,bylinskiiWhatDifferentEvaluation2018}, making progress difficult to assess.

To address this, a probabilistic evaluation framework was proposed~\cite{kummererInformationtheoreticModelComparison2015,kummererPredictingVisualFixations2023} in which fixation prediction is cast as density estimation (see also \cite{baddeleyHighFrequencyEdges2006,barthelmeModelingFixationLocations2013}) and model performance is measured in information-theoretic units (bits per fixation). This framework motivates cross-validated bandwidth optimization and naturally accommodates mixture components such as center bias and uniform regularization.%

Mixture models provide a natural framework for incorporating prior knowledge into empirical density estimation. Center bias---the tendency to fixate near the image center---is well documented~\cite{tatlerCentralFixationBias2007,clarkeDerivingAppropriateBaseline2014} and has been modeled as a separate mixture component in fixation prediction~\cite{kummererInformationtheoreticModelComparison2015,kummererPredictingVisualFixations2023}. K\"ummerer and Bethge~\cite{kummererPredictingVisualFixations2023} introduced a three-component mixture (KDE, center bias, and uniform) with cross-validated bandwidth optimization. K\"ummerer et al.~\cite{kummererModelingSaliencyDataset2025} extended this with dataset-level bias parameters and per-image optimization of the mixture weights; despite this per-image optimization, state-of-the-art models still exceed the resulting consistency estimates. D'Agostino et al.~\cite{dagostinoWhatMovesEyes2025} subsequently included a saliency model component in the mixture, but only as input for a scanpath model, not for density estimation.

\subsubsection{Interobserver Consistency and Upper Bounds}
\label{sec:related:consistency}
Interobserver consistency---how well one observer's fixations can be predicted from those of others---provides a natural ceiling for fixation prediction models. Wilming et al.~\cite{wilmingMeasuresLimitsModels2011} derived lower and upper bounds for saliency metrics and showed that approximately 20 observers suffice for a stable AUC upper bound. Judd et al.~\cite{juddBenchmarkComputationalModels2012} used a single held-out observer to estimate performance ceilings on the MIT300 benchmark.
Koehler et al.~\cite{koehlerWhatSaliencyModels2014} provided per-category consistency estimates, revealing substantial variation across image types.
Engelke et al. \cite{engelkeComparativeStudyFixation2013} analysed consistency across different laboratories, and
more recently, Wu et al.~\cite{wuEvaluatingCrossSubjectCrossDevice2025} examined cross-subject and cross-device consistency, finding that individual-to-average consistency varies substantially across observers and devices. Recently,~\cite{kummererModelingSaliencyDataset2025} showed that state-of-the-art models can exceed classical consistency estimates, hinting at problems with current ceiling estimation methodology.
De Haas et al.~\cite{haasIndividualDifferencesVisual2019} showed that interobserver variability is structured along semantic dimensions (e.g., faces, text) rather than being pure noise, suggesting that the gap between individual observers and the average density contains systematic signal that better estimation methods might capture.

\subsubsection{Adaptive Bandwidth Methods}
\label{sec:related:adaptive}

Adaptive bandwidth methods address a fundamental limitation of fixed-bandwidth KDE: a single bandwidth cannot simultaneously resolve fine structure in dense regions and avoid noise amplification in sparse regions. Early work on nearest-neighbor methods~\cite{loftsgaardenNonparametricEstimateMultivariate1965a,breimanVariableKernelEstimates1977} determined bandwidth from local data spacing, effectively adapting to local density. Abramson~\cite{abramsonBandwidthVariationKernel1982} subsequently proposed scaling the bandwidth at each data point inversely with the square root of a pilot density estimate, yielding a simple and effective adaptive estimator. Despite their widespread use in spatial statistics, adaptive bandwidth methods have not previously been applied to fixation density estimation. Our approach combines adaptive bandwidth estimation with the mixture model framework reviewed above.

\section{Methods}
\label{sec:methods}

\subsection{Background: Probabilistic Framework}
\label{sec:methods:background}

We formulate fixation prediction as a density estimation problem following K\"ummerer et al.~\cite{kummererInformationtheoreticModelComparison2015}. Given an image~$I$, we seek a probability density $p(\mathbf{x} \mid I)$ over spatial locations $\mathbf{x} = (x, y)$ that assigns high probability to locations where observers fixate. Model quality is measured by the average log-likelihood assigned to held-out fixations,
\begin{equation}
    \label{eq:loglik}
    \mathcal{L} = \frac{1}{N}\sum_{i=1}^{N} \log p(\mathbf{x}_i \mid I)\,,
\end{equation}
where $\{\mathbf{x}_i\}_{i=1}^{N}$ are fixation locations not used to fit~$p$. This information-theoretic evaluation avoids the pathologies of map-comparison metrics and provides a principled, calibration-free measure in units of bits per fixation (after converting to $\log_2$, an usually reported relative to a baseline model).

The standard approach to constructing an empirical fixation density is Gaussian kernel density estimation. Given fixations $\{\mathbf{x}_j\}_{j=1}^{M}$ from $M$ observers on a single image, the fixed-bandwidth Gaussian KDE is
\begin{equation}
    \label{eq:kde_fixed}
    \hat{p}_{\mathrm{KDE}}(\mathbf{x}) = \frac{1}{M}\sum_{j=1}^{M} \mathcal{N}(\mathbf{x};\, \mathbf{x}_j, h^2 \mathbf{I})\,,
\end{equation}
where $h$ is a scalar bandwidth and~$\mathbf{I}$ the $2\times 2$ identity matrix. In practice, $h$ is typically set to a fixed constant (often one degree of visual angle) and held constant across images. Evaluating Eq.~\ref{eq:loglik} via leave-one-subject-out (LOSO) cross-validation---training on fixations from all but one observer and scoring on the held-out observer---yields the \emph{interobserver consistency} (IOC), which serves as the ceiling against which saliency models are compared.

\subsection{Mixture Model for Empirical Fixation Densities}
\label{sec:methods:mixture}

A pure KDE must explain every fixation from the Gaussian kernels placed at other observers' fixation locations alone. With only ${\sim}15$ observers per image (a typical number in eye-tracking datasets \cite{juddLearningPredictWhere2009,chenCharacterizingTargetAbsentHuman2022}), this may be insufficient: some fixations fall in the tails of all training kernels, receiving very low probability. In particular, there is a set of fixation patterns that are common across \emph{all} images---most prominently the well-documented center bias~\cite{tatlerCentralFixationBias2007,clarkeDerivingAppropriateBaseline2014}---that would need to be learned anew by the KDE for every image.

We address this by embedding the KDE in a mixture model. For an image~$I$ with training fixations $\{\mathbf{x}_j\}$, the empirical density is
\begin{equation}
    \label{eq:mixture}
    p(\mathbf{x} \mid I) = w_{\mathrm{KDE}}\, \hat{p}_{\mathrm{KDE}}(\mathbf{x}) \;+\; w_{\mathrm{CB}}\, p_{\mathrm{CB}}(\mathbf{x}) \;+\; w_{\mathrm{U}}\, p_{\mathrm{U}}(\mathbf{x}) \;+\; w_{\mathrm{S}}\, p_{\mathrm{S}}(\mathbf{x} \mid I)\,,
\end{equation}
where the weights satisfy $\sum_k w_k = 1$, $w_k \geq 0$.
Here $\hat{p}_{\mathrm{KDE}}$ is the kernel density estimate from other observers' fixations on the same image, $p_{\mathrm{CB}}$ a center-bias prior estimated from all other images~\cite{kummererInformationtheoreticModelComparison2015}, $p_{\mathrm{U}} = 1/(W \cdot H)$ a uniform regularization floor, and $p_{\mathrm{S}}$ the density from a state-of-the-art saliency model (DeepGaze MSDB~\cite{kummererModelingSaliencyDataset2025}).
The KDE captures image-specific spatial structure (where do other observers look in this image), the center bias captures general spatial tendencies (where do people look in other images), and the saliency model captures general semantic tendencies (what image features do people look at in other images). The uniform component accounts for noise and idiosyncratic fixations. Together, they regularize the image-specific KDE with prior information that a small number of observers cannot provide on their own.

\subsection{Adaptive Bandwidth KDE}
\label{sec:methods:adaptive}

\begin{wrapfigure}{r}{0.52\textwidth}
    \centering
    \vspace{-0.9cm}
    \includegraphics[]{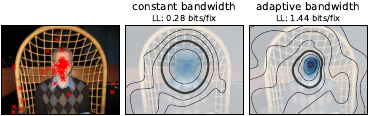}
    \caption{Adaptive bandwidth KDE. Left: classic fixed-bandwidth Gaussian KDE. Right: Abramson adaptive bandwidth KDE, where each source fixation contributes with a different bandwidth determined by the pilot density estimate. Where fixations cluster, bandwidths shrink to preserve spatial detail; where fixations are sparse, bandwidths grow to remain smooth. In both cases, parameters are optimized for leave-one-subject-out prediction.
    }
    \label{fig:abramson}
    \vspace{-0.7cm}
\end{wrapfigure}

A fixed bandwidth~$h$ imposes the same spatial resolution everywhere: in regions where fixations cluster tightly (e.g., around a salient face), the kernels are too broad and blur genuine structure; in sparse regions (e.g., background), the same bandwidth may overfit to individual fixation locations. With only ${\sim}15$ observers per image, this trade-off is especially acute.

Abramson~\cite{abramsonBandwidthVariationKernel1982} proposed a simple two-stage adaptive estimator. In the \emph{first stage}, a pilot density~$\tilde{p}$ is estimated. Here, we use a standard fixed-bandwidth KDE with pilot bandwidth~$h_0$.
In the \emph{second stage}, each data point receives its own bandwidth inversely proportional to the square root of the pilot density evaluated at that point:
    $h_j = \frac{\alpha}{\sqrt{\tilde{p}(\mathbf{x}_j)}}$,
where $\alpha$ is a proportionality constant. The adaptive KDE is then
    $\hat{p}_{\mathrm{KDE}}(\mathbf{x}) = \frac{1}{M}\sum_{j=1}^{M} \mathcal{N}(\mathbf{x};\, \mathbf{x}_j, h_j^2\,\mathbf{I})$.
Where fixations are dense, the pilot density is high and bandwidths shrink, preserving fine-grained spatial structure. Where fixations are sparse, the pilot density is low and bandwidths grow, allowing the estimate to remain smooth rather than over-fitting to isolated points (see Figure~\ref{fig:abramson}).

\subsection{Parameter Optimization}
\label{sec:methods:optimization}

The full model has the following free parameters: for the KDE component, either a single bandwidth~$h$ (fixed-bandwidth KDE) or the pilot bandwidth~$h_0$ and scaling constant~$\alpha$ (adaptive KDE); and for the mixture, one log-weight per regularizer component (center bias, uniform, saliency model). All parameters are optimized jointly per image by maximizing the LOSO cross-validated log-likelihood (Eq.~\ref{eq:loglik}). The entire computation graph is implemented in PyTorch and is fully differentiable, providing exact gradients and enabling efficient per-image optimization. Implementation details are provided in Appendix~\ref{app:optimization}.

\subsection{LOSO Density Estimation}
\label{sec:methods:loso}

\begin{figure}[tb]
    \centering
    \includegraphics[]{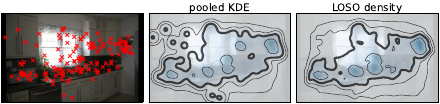}
    \caption{Locally crossvalidated density. Pooling all fixations into a single KDE (left) overfits to individual fixation positions. Our locally weighted averaging of LOSO models (right) constructs a spatial density where each location is dominated by the fold that excluded the nearest observer, providing a more faithful density representation of interobserver consistency.}
    \label{fig:loso}
\end{figure}

Many downstream uses require a single spatial density map per image---for visualization, map-comparison metrics, or scanpath models. Simply pooling all fixations into one KDE introduces a subtle overfitting problem: each fixation has a kernel centered at its own location, inflating density at observed positions (Figure~\ref{fig:loso}, left). We instead construct a \emph{locally crossvalidated} density via weighted averaging of LOSO models.
The key property is that at each fixation location, the density is dominated by the LOSO model that \emph{excluded} that fixation's observer: precisely the model used during cross-validated evaluation (Figure~\ref{fig:loso}, right). Details are in Appendix~\ref{app:loso}. All visualizations in this paper use the locally crossvalidated density.

\subsection{Density Visualization}
\label{sec:methods:visualization}

Fixation densities are most commonly visualized as heatmaps overlaid on the stimulus image \cite{bylinskiiWhereShouldSaliency2016}. However, standard approaches have systematic shortcomings: per-image normalization makes cross-image comparison impossible (a nearly uniform distribution and a highly peaked one can appear equally ``hot''), while shared color scales hide weaker structure and still provide only coarse qualitative impressions, obscuring quantitative structure such as the relative magnitude of secondary peaks.

We propose a visualization that addresses both issues by combining a \emph{saturating heatmap} with \emph{log-spaced contour lines}. The heatmap expresses density as a ratio to the uniform density and applies a saturating nonlinearity, so that near-uniform distributions appear pale while peaked distributions saturate---enabling immediate cross-image comparison of overall ``peakedness'' without per-image rescaling. The contour lines are placed at density levels that are integer powers of a base~$\gamma$ relative to the uniform density, with a thicker line at the uniform level. Each contour crossing thus represents a fixed multiplicative change in density, providing quantitative reference points that faithfully reflect the actual dynamic range---unlike quantile-based contours \cite{kummererMatthiaskPysaliency2026,kummererSaliencyBenchmarkingMade2018}, which can suggest strong spatial structure even in nearly uniform distributions. The exact formulas and parameter choices are given in Appendix~\ref{app:visualization}.

Figure~\ref{fig:densityvisualization} compares our method against the mentioned alternatives (per-image heatmaps, shared-scale heatmaps, and quantile contours) on images spanning a range of peakedness. Our method avoids the pitfalls of each: it neither fabricates structure in uniform distributions nor hides weaker structure, and enables both qualitative and quantitative comparisons across and within images.

\begin{figure}[tb]
    \centering
    \includegraphics[]{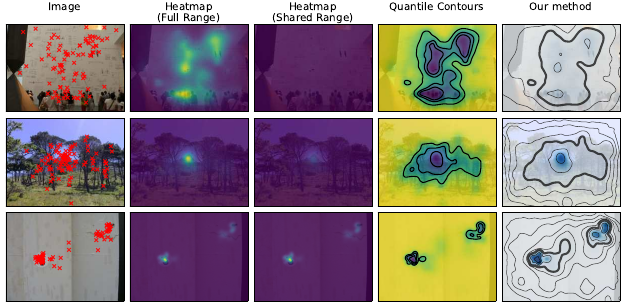}
    \caption{Comparison of density visualization methods on three images with increasing peakedness.
    For each image, we visualize our empirical density with four different methods: Per-image heatmaps (first column) suggest strong patterns even in near-uniform distributions. Shared-scale heatmaps \cite{bruceSaliencyAttentionVisual2009} (second column) enable comparison but obscure detail. Quantile contours \cite{kummererMatthiaskPysaliency2026,kummererSaliencyBenchmarkingMade2018,linardosDeepGazeIIECalibrated2021,kummererModelingSaliencyDataset2025} (third column) overemphasize structure in uniform distributions and suppress secondary peaks in peaked ones. Our method (fourth column) combines a saturating heatmap anchored at uniform density with log-spaced contour lines, providing both immediate qualitative comparison across images and quantitative spatial detail within each image.}
    \label{fig:densityvisualization}
\end{figure}

\section{Experiments}
\label{sec:experiments}

\subsection{Experimental Setup}
\label{sec:experiments:setup}

We evaluate our method on four datasets spanning a range of observer counts: MIT1003~\cite{juddLearningPredictWhere2009} (1003 images, 15 subjects), CAT2000 train split~\cite{borjiCAT2000LargeScale2015} (2000 images, 18 subjects), COCO-Freeview validation split~\cite{chenCharacterizingTargetAbsentHuman2022,yangPredictingHumanAttention2023} (603 images, 10 subjects), and DAEMONS validation split~\cite{schwetlickPotsdamDataSet2024} (200 images, 200 subjects). We additionally report results on the MIT300~\cite{juddBenchmarkComputationalModels2012}, CAT2000 and COCO-Freeview test holdout sets from the MIT/T\"ubingen Saliency Benchmark~\cite{kummererMITTuebingenSaliency}. We remove the artifact reported by \cite{kummererModelingSaliencyDataset2025} from the CAT2000 dataset (see also Appendix~\ref{app:datasets}).

Our primary metric is log-likelihood (LL), measured in bits per fixation relative to a uniform basline model \cite{kummererInformationtheoreticModelComparison2015}. Log-likelihood is information-theoretic, unbounded above, and---unlike AUC---does not saturate as models improve, making it ideal for detecting differences between high-performing density estimates. We report AUC as secondary metric for comparability with prior work.

The \emph{baseline} follows the current MIT/T\"ubingen Saliency Benchmark protocol: a fixed-bandwidth Gaussian KDE (Eq.~\ref{eq:kde_fixed}) mixed with uniform and center bias components, with all parameters optimized jointly across the dataset. The \emph{full model} uses the mixture defined in Eq.~\ref{eq:mixture}: adaptive-bandwidth KDE via Abramson's method~(Sec.~\ref{sec:methods:adaptive}), uniform and center bias components, and DeepGaze MSDB~\cite{kummererModelingSaliencyDataset2025} as a saliency regularizer. DeepGaze MSDB was trained jointly on all four development datasets plus FIGRIM, making it well-suited as a cross-image semantic prior; for each image, we use a checkpoint trained without ever seeing that image to ensure proper cross-validation of the saliency component.
All parameters are optimized per image via leave-one-subject-out (LOSO) cross-validation~(Sec.~\ref{sec:methods:optimization}), yielding image-specific density estimates. To trace where improvements originate, we also run differently ablated model variants, including the still commonly used fixed-bandwidth KDE with one degree of visual angle.

We extend our analysis to video with the LEDOV video saliency dataset~\cite{jiangDeepVSDeepLearning2018} (538 videos, 32 subjects). For the saliency model component in this setting, in place of DeepGaze MSDB, we use UniformerSal~\cite{el-jizIsolatingRoleTemporal}, one of the best-performing models on this dataset.

\begin{figure}[tb]
    \centering
    \noindent
    \begin{tikzpicture}
        \coordinate (hsep) at (-0.3, 0);
        \coordinate (vsep) at (0, -5);
        \coordinate (labelsep) at (-0.5, -0.0);
        \tikzset{label/.style={font=\sffamily\bfseries}}
        \tikzset{anchor=north west}

        \node (IG_MIT) at (0, 0) {\includegraphics[]{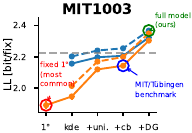}};
        \node (IG_CAT) at ($ (IG_MIT.north east) + (hsep) $) {\includegraphics[]{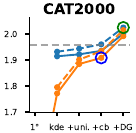}};
        \node (IG_COCO) at ($ (IG_CAT.north east) + (hsep) $) {\includegraphics[]{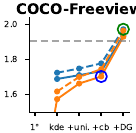}};
        \node (IG_DAE) at ($ (IG_COCO.north east) + (hsep) $) {\includegraphics[]{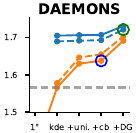}};
        \node (legend) at ($ (IG_DAE.north east) + (hsep) + (0.08, 0) $) {\includegraphics[]{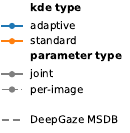}};
    \end{tikzpicture}

    \caption{Interobserver Consistency Estimates: For four datasets, we compare estimates of inter-observer consistency (IOC) using different variants of our proposed model. We vary the KDE type (classic vs.\ Abramson adaptive bandwidth), the optimization procedure (global vs.\ per-image), and the mixture components (KDE only, KDE with uniform and center bias, and full model with additional DeepGaze MSDB component). For comparison we also add the commonly used Gaussian KDE with a fixed bandwidth of one degree of visual angle, which is not optimized for the data.}
    \label{fig:main_experiment}
    \vspace{-0.5cm}
\end{figure}
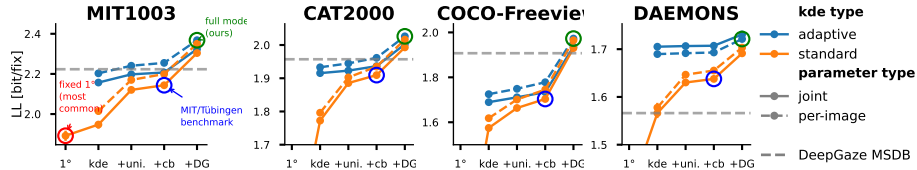

\subsection{Results}
\label{sec:experiments:results}

\subsubsection{Interobserver consistency is substantially higher than previously estimated.}
\label{sec:experiments:main}
Figure~\ref{fig:main_experiment} traces the progression from baseline density estimation to our full model across four datasets. In all cases, the full model yields substantially higher IOC estimates---gains range from approximately 0.15 to 0.4~bit/fix depending on dataset characteristics. These improvements are confirmed on held-out benchmark data (Table~\ref{tab:benchmark}, Figure~\ref{fig:teaser}), where we observe consistent gains in both information-theoretic and AUC-based metrics.

Notably, the still commonly used fixed-bandwidth KDE with one degree of visual angle performs so poorly that we truncate it in most panels to preserve visibility of other effects---a clear indication that this practice should be abandoned in favor of optimized alternatives.

The improved estimates also clarify the gap between current models and human consistency: under the old methodology, IOC appeared below DeepGaze MSDB on nearly all datasets, but our new estimates restore a meaningful ceiling, indicating that substantial room for model improvement remains.

\begin{table*}[t]
    \centering
    \small
    \caption{Interobserver consistency estimates on the MIT/T\"ubingen Saliency Benchmark holdout sets. We compare the old method used in the Benchmark so far against our new method, and include the current SOTA model on all three datasets, DeepGaze MSDB. We find substantially higher IOC estimates on all datasets}%
    \label{tab:benchmark}
    \begin{tabular}{l@{\hskip 0.2in}r@{\hskip 0.1in}r@{\hskip 0.2in}r@{\hskip 0.1in}rr@{\hskip 0.0in}r}
    \toprule
    dataset & \multicolumn{2}{c}{MIT300} & \multicolumn{2}{r}{CAT2000} & \multicolumn{2}{r}{COCO-Freeview} \\
    \cmidrule(lr){2-3}\cmidrule(lr){4-5}\cmidrule(lr){6-7}
    & IG & AUC & IG & AUC & \qquad IG & AUC \\
    model &  &  &  &  &  &  \\
    \midrule
    DeepGaze MSDB & 1.26 & 89.4\% & 0.52 & 88.7\% & 1.07 & 89.7\% \\
old IOC estimate & 1.34 & 89.9\% & 0.49 & 88.5\% & 0.87 & 88.3\% \\
new IOC estimate & 1.45 & 90.5\% & 0.59 & 89.1\% & 1.14 & 90.0\% \\

    \bottomrule
    \end{tabular}
\end{table*}

\begin{table*}[b]
    \centering
    \small
    \caption{Per-image improvement quantiles for the full model over the baseline model across datasets. We report improvements as differences in LL and AUC, and for LL additionally relative improvements compared to the baseline performance.}
    \label{tab:improvement_quantiles_joint}
    \begin{tabular}{l *{12}{r}}
    \toprule
    & \multicolumn{3}{c}{MIT1003} & \multicolumn{3}{c}{CAT2000} & \multicolumn{3}{c}{COCO-Freeview} & \multicolumn{3}{c}{DAEMONS} \\
    \cmidrule(lr){2-4}\cmidrule(lr){5-7}\cmidrule(lr){8-10}\cmidrule(lr){11-13}
    Quantile & LL & LL$_{\mathrm{rel}}$ & AUC & LL & LL$_{\mathrm{rel}}$ & AUC & LL & LL$_{\mathrm{rel}}$ & AUC & LL & LL$_{\mathrm{rel}}$ & AUC \\
    \midrule
    50\% & 0.21 & 0.10 & 1.1\% & 0.11 & 0.06 & 0.7\% & 0.26 & 0.15 & 1.7\% & 0.08 & 0.05 & 0.7\% \\
75\% & 0.28 & 0.14 & 1.6\% & 0.14 & 0.08 & 1.0\% & 0.33 & 0.20 & 2.3\% & 0.10 & 0.07 & 0.9\% \\
95\% & 0.39 & 0.26 & 2.7\% & 0.20 & 0.12 & 1.6\% & 0.43 & 0.35 & 3.5\% & 0.13 & 0.13 & 1.4\% \\
99\% & 0.54 & 0.40 & 3.8\% & 0.23 & 0.16 & 2.1\% & 0.50 & 0.51 & 4.7\% & 0.14 & 0.15 & 1.7\% \\
100\% & 0.73 & 2.23 & 5.3\% & 0.33 & 0.32 & 3.6\% & 0.68 & 0.85 & 5.5\% & 0.17 & 0.20 & 2.0\% \\

    \bottomrule
    \end{tabular}
\end{table*}

\subsubsection{Individual images can gain far more than the dataset average.}
\label{sec:experiments:perimage}
\begin{figure}[tb]
    \centering
    \includegraphics[]{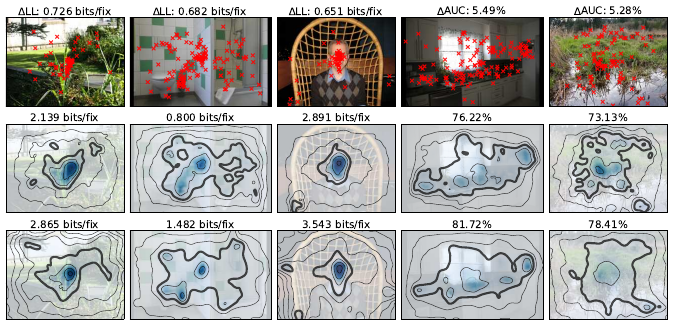}
    \caption{Images where LOSO prediction improves most from baseline to full model in terms of LL or AUC.}
    \label{fig:casestudy_mostimproved}
    \vspace{-0.5cm}
\end{figure}

Table~\ref{tab:improvement_quantiles_joint} reveals that per-image improvements substantially exceed the dataset averages. While median gains are already meaningful (\MITMedianLLGain--\COCOMedianLLGain~bit/fix in LL, \MITMedianAUCGain--\COCOMedianAUCGain{} percentage points in AUC), the 95th and 99th percentiles reach 2--3$\times$ these values. Maximum improvements approach \MITMaxLLGain~bit/fix in LL and exceed \MITMaxAUCGain\% in AUC---for context, the entire gap between center bias and state-of-the-art models on MIT1003 is approximately 1.2~bit/fix. Figure~\ref{fig:casestudy_mostimproved} illustrates images achieving the largest gains from baseline to our full model. Notably, LL and AUC improvements are driven by complementary failure modes: LL gains are largest on images with tightly clustered fixations, where fixed-bandwidth kernels blur fine structure, while AUC gains dominate on images with dispersed fixations, where the baseline overfits to sparse samples. This dichotomy has direct implications for sample-level analyses such as failure case studies or inverse benchmarking: the old density estimates were most inadequate precisely for the images where accurate estimation matters most.

\begin{figure}[tb]
    \centering
    \includegraphics[]{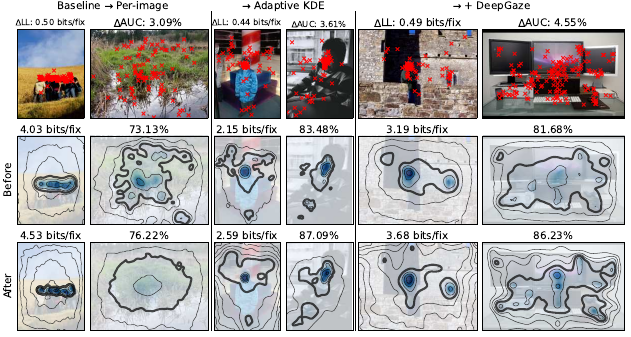}
    \caption{Per-component case studies. Starting with a classic Gaussian KDE with uniform and centerbias mixture components and parameters optimized on the full dataset, we (1) change to parameters optimized per image, (2) switch to Abramson adaptive bandwidth KDE, and (3) add the DeepGaze MSDB mixture component. For each step, we show the two images where performance improves most in terms of LL and AUC. See Appendix~\ref{app:per-image-analysis} for extended per-image analyses for each step, also including even simpler model variants.}
    \label{fig:casestudy_ablations}
\end{figure}

\subsubsection{Adaptive bandwidth and saliency regularization are the main drivers.}
\label{sec:experiments:ablation}
Figure~\ref{fig:main_experiment} also includes the commonly used fixed one-degree bandwidth KDE (often cut off due to substantially worse performance) and intermediate ablation variants.
Adaptive KDE alone---without any regularizers---already yields substantial gains across all datasets. By adapting bandwidths to local data density, it provides implicit regularization that captures multi-scale spatial structure. Notably, this holds even on DAEMONS (200 subjects/image), indicating the benefit stems from fundamentally better density estimation rather than merely compensating for sparse data. The saliency regularizer (DeepGaze MSDB) contributes significant additional gains, but its magnitude is dataset-dependent: large on COCO-Freeview (10 subjects/image), modest on DAEMONS. This clean interaction---fewer observers yield more benefit from cross-image semantic priors---confirms that the regularizer fills gaps left by limited fixation counts. Per-image bandwidth optimization provides the smallest aggregate contribution but remains important for specific images.

Figure~\ref{fig:casestudy_ablations} reveals that each component rescues a distinct failure mode. Per-image optimization shows the largest effect on images where object size deviates strongly from the global average bandwidth---small faces requiring narrow kernels, or global scene layouts requiring broad ones. Adaptive KDE contributes most when small and large structures co-occur within the same image, where no single bandwidth suffices. The saliency regularizer provides the greatest benefit for small objects with too few fixations to estimate shape, or for images dense with many interesting objects where per-object fixation counts are low. Critically, all three components yield gains of similar magnitude on their respective most-affected images ($\sim$0.5~bit/fix in LL, 3--5\% in AUC).
See  Appendix~\ref{app:per-image-analysis} for extended per-image analyses for each step.

\subsubsection{Cross-validation design matters: LOSO and LOFO yield different consistency estimates.}
\label{sec:experiments:loso_vs_lofo}
All results above use leave-one-subject-out (LOSO) cross-validation, which estimates how well fixations from a new, previously unseen observer can be predicted. An alternative is leave-one-fixation-out (LOFO) cross-validation, which holds out individual fixations rather than entire subjects. LOFO estimates how well a new fixation from an \emph{unknown} subject within a \emph{known} pool of observers can be predicted---it conditions on knowledge of the subject population without knowing which subject produced a given fixation. Which design is appropriate depends on the benchmark: MIT300, where test fixations come from genuinely different observers recorded years later~\cite{juddBenchmarkComputationalModels2012}, calls for LOSO; benchmarks where the same subjects appear in both training and test sets are better served by LOFO. We use LOSO throughout this paper because it answers the more practically relevant question ``how well can we predict a new observer?'', which is the setting faced by most applications and also the standard interpretation of interobserver consistency. However, all of our proposed improvements apply equally to LOFO estimation.

\begin{figure}[tb]
    \centering
    \begin{tikzpicture}
        \coordinate (hsep) at (-0.3, 0);
        \coordinate (labelsep) at (0.2, -0.5);
        \tikzset{label/.style={font=\sffamily\bfseries}}
        \tikzset{anchor=north west}

        \node (panelA) at (0, 0) {\includegraphics[width=0.45\textwidth]{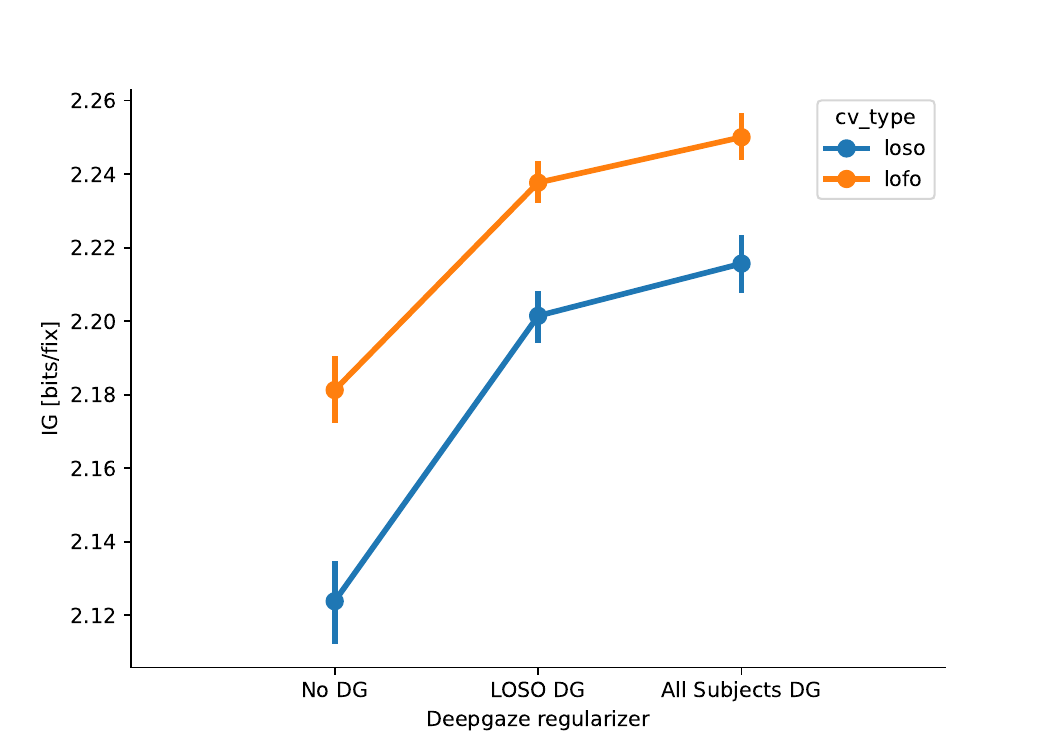}};
        \node (panelB) at ($ (panelA.north east) + (hsep) $) {\includegraphics[width=0.45\textwidth]{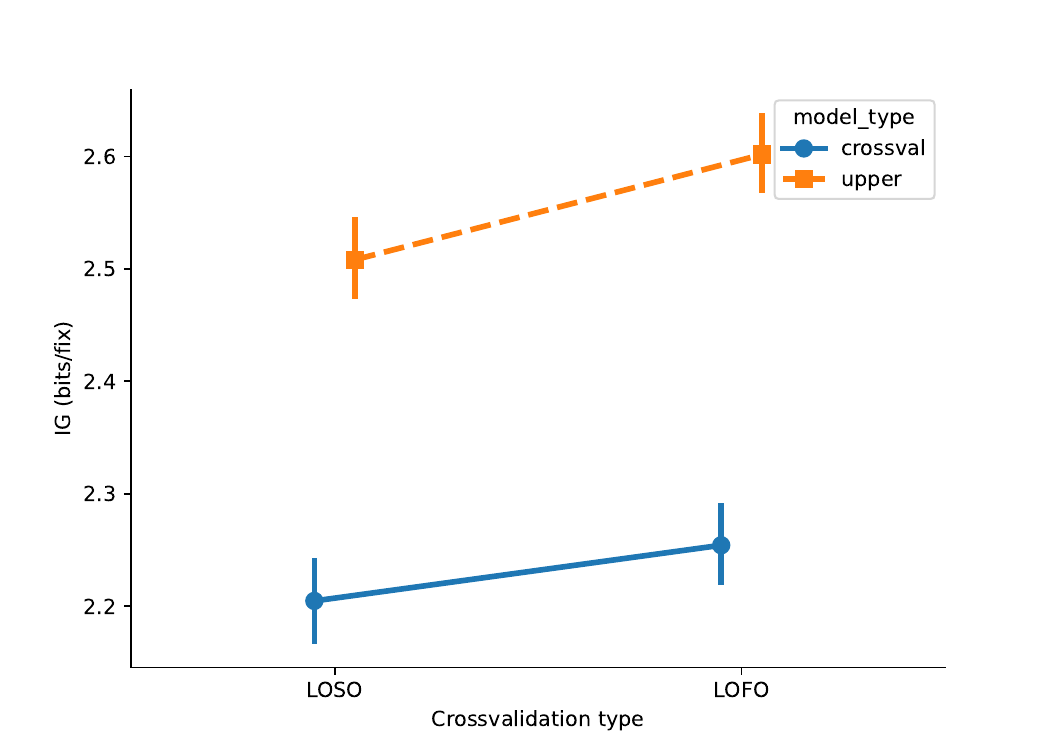}};
        \node[label, anchor=south] at ($ (panelA.north west) + (labelsep) $) {(a)};
        \node[label, anchor=south] at ($ (panelB.north west) + (labelsep) $) {(b)};
    \end{tikzpicture}
    \caption{Cross-validation design for interobserver consistency estimation on MIT1003. (a)~LOSO vs.\ LOFO cross-validation for our full model with different DeepGaze configurations. LOFO consistently yields higher consistency estimates, reflecting its access to subject pool information. The difference between the standard DeepGaze MSDB and a LOSO-trained variant is small, justifying the use of the standard model. (b)~Crossvalidated vs.\ pooled (``upper'') density estimation. Pooled estimates, which include each fixation's own kernel, overfit by \LFUpperOverfitLOSO{}--\LFUpperOverfitLOFO{}~bit/fix compared to proper cross-validation. The LOSO--LOFO difference (small within each group) is dwarfed by the crossvalidation--pooling gap.}
    \label{fig:cv_design}
\end{figure}
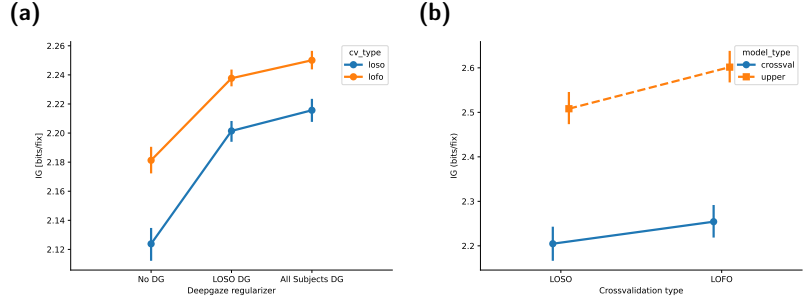

Figure~\ref{fig:cv_design}a compares LOSO and LOFO on a crossvalidation fold of MIT1003 (\LFNStimuli{} images) for our full model across three DeepGaze configurations: no saliency model, a DeepGaze variant trained in a matched LOSO fashion (excluding the test subject's data from DeepGaze training), and the standard DeepGaze MSDB. LOFO consistently yields higher consistency estimates than LOSO (\LFLOFOCrossval{} vs.\ \LFLOSOCrossval{}~bit/fix with standard DeepGaze, a gap of \LFGapLOSOLOFO{}~bit/fix), as expected: conditioning on the known subject pool provides strictly more information. The gap is largest without any saliency model component (\LFGapLOSOLOFONoDG{}~bit/fix), suggesting that DeepGaze already captures some between-subject variation.

A secondary concern is that our standard DeepGaze MSDB was trained on data including the test subjects' fixations, potentially leaking subject-specific information into the LOSO evaluation. Comparing against the LOSO-trained DeepGaze variant reveals a difference of only \LFDGLeakage{}~bit/fix, confirming that this leakage is minor. We therefore use the standard DeepGaze MSDB throughout, accepting this small methodological impurity to avoid the substantial computational cost of training separate LOSO DeepGaze models for each fold and dataset.

\subsubsection{Pooled fixation densities massively overfit and should be replaced by crossvalidated estimates.}
\label{sec:experiments:upper_overfitting}
A common practice in saliency benchmarking is to estimate interobserver consistency by pooling all observers' fixations into a single density without cross-validation of the KDE---the approach currently used for the ``upper gold standard'' on the MIT/T\"ubingen Saliency Benchmark~\cite{kummererMITTuebingenSaliency}. Unlike proper cross-validation, this pooled estimate includes each fixation's own kernel in the density used to evaluate it, inflating scores at observed fixation positions.

Figure~\ref{fig:cv_design}b quantifies the resulting overfitting. Pooled estimates exceed their properly crossvalidated counterparts by \LFUpperOverfitLOSO{}--\LFUpperOverfitLOFO{}~bit/fix---an overestimation comparable in magnitude to the entire improvement from baseline to full model reported in Section~\ref{sec:experiments:main}. We therefore recommend that benchmarks replace pooled ``upper'' gold standards with properly crossvalidated estimates: LOFO as the primary within-dataset ceiling (since most benchmarks share subjects between training and test), supplemented by LOSO to characterize generalization to new observers.

An extended analysis across all DeepGaze configurations (Appendix~\ref{app:cv_design}, Figure~\ref{app:fig:crossval_vs_upper_full}) reveals that the overfitting is most severe without a saliency model component (\LFUpperOverfitNoDGLOSO{}~bit/fix for LOSO) and, strikingly, that the ordering between DeepGaze variants \emph{reverses} between crossvalidated and pooled evaluation: adding DeepGaze improves crossvalidated estimates but \emph{decreases} pooled scores, because without DeepGaze the KDE receives all mixture weight and the self-prediction artifact is amplified. This reversal confirms that pooled estimates reflect the degree of self-prediction rather than genuine predictive performance.

\begin{figure}[tb]
    \centering
    \includegraphics[]{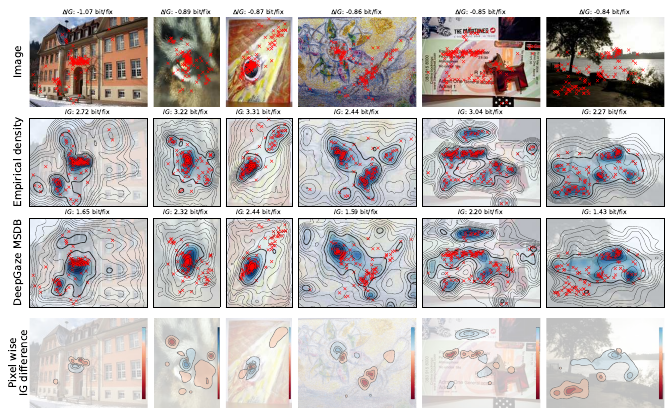}
    \caption{Remaining SOTA model failures. We show the six images where DeepGaze MSDB~\cite{kummererModelingSaliencyDataset2025} loses most performance compared to our improved empirical density, in terms of LL. Pixel-space information gain~\cite{kummererInformationtheoreticModelComparison2015} reveals precisely where model predictions are most discrepant from the empirical density. We find clear failure categories such as paintings, context-dependent saliency, and semantic text importance.}
    \label{fig:casestudy_deepgaze}

\end{figure}

\subsubsection{Better density estimates reveal previously hidden SOTA model failures.}
\label{sec:experiments:casestudy_sota}
To demonstrate the practical value of improved density estimation, we compare per-image performance of DeepGaze MSDB~\cite{kummererModelingSaliencyDataset2025}, the current state-of-the-art saliency model, against our improved IOC estimate. Figure~\ref{fig:casestudy_deepgaze} shows the six images where DeepGaze MSDB loses most log-likelihood; pixel-wise information gain maps~\cite{kummererInformationtheoreticModelComparison2015} reveal precisely where predictions diverge from the empirical density in data-relevant locations. Three failure categories emerge. First, two of the six worst images are paintings.
Second, the model struggles with context-dependent saliency: objects that are not inherently salient but become so due to scene composition, such as road markings in the foreground of a river scene. Third, DeepGaze predicts visually prominent text (large fonts, special styling) rather than semantically important text (e.g., an artist name on a small ticket). The quality of the density estimate determines which failures are visible: the original DeepGaze MSDB paper~\cite{kummererModelingSaliencyDataset2025} performed a similar failure analysis using the old IOC but identified \emph{different} failure images with much smaller errors (0.2--0.35~bit/fix vs.\ 0.84--1.0~bit/fix here). Improved density estimation increases the signal-to-noise ratio of per-image analysis, allowing genuine model failures to surface more reliably rather than being obscured by noise in the gold standard itself. These results indicate substantial remaining headroom for model improvement, with concrete directions: better handling of unusual image types, context-dependent saliency, and semantic text importance.

\begin{wrapfigure}{r}{0.52\textwidth}
    \centering
    \vspace{-0.9cm}
    \includegraphics[]{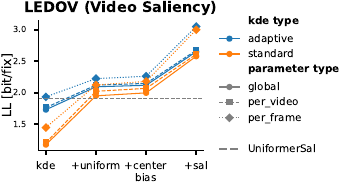}
    \caption{Extension to video: IOC estimates on LEDOV for different model variants and optimization granularities. Our full model reveals substantial headroom beyond what standard methodology suggested.}
    \label{fig:video_extension}
    \vspace{-0.5cm}
\end{wrapfigure}

\subsubsection{Video saliency is far from solved.}
\label{sec:experiments:video}
We extend our framework to video saliency on LEDOV (Figure~\ref{fig:video_extension}).
In the video setting, only $\sim$30ms of gaze data is available per subject and frame, making density estimation particularly challenging.
Under standard methodology, UniformerSal (\VideoUniformerSalLL{}~bit/fix) nearly matches IOC (\VideoOldIOC{}~bit/fix), suggesting the benchmark is solved. Our framework revises this dramatically: IOC rises to \VideoNewIOC{}~bit/fix, revealing that SOTA leaves more than half the explainable information unexplained. In this low-data regime, the video saliency model provides semantic and dynamic scaffolding while adaptive bandwidths prevents oversmoothing or overfitting.

\section{Discussion and Conclusion}
\label{sec:discussion}

Our results demonstrate that interobserver consistency---the ceiling against which saliency models are measured---is not a fixed property of a dataset but an evolving estimate that depends critically on the quality of density estimation.
As estimation methods improve, the ceiling rises, benchmark headroom increases, and sample-level analyses become more reliable. This has a practical implication: benchmarks should be periodically re-evaluated with improved estimation methods rather than treated as static. We will make the code of our method available for easy application. %

\paragraph{Modularity and implications for model development.}
Our framework is modular: substantial gains accrue even without a saliency model component. This matters because strong saliency models exist mainly for natural scene free-viewing; in applied domains such as radiology, pathology, or driving, no comparably effective model is available. Where no suitable model exists, adaptive-bandwidth KDE with per-image optimization becomes the critical improvement. Beyond raising the ceiling, improved density estimates enable sharper failure case analysis: the images where DeepGaze MSDB underperforms (Fig.~\ref{fig:casestudy_deepgaze}) reveal concrete failure modes---context-dependent saliency, paintings, semantic text---that provide actionable directions for continued model development.

\paragraph{Practical recommendations.}
At minimum, bandwidth should be optimized via leave-one-subject-out cross-validation rather than set to a fixed default---this costs almost nothing and always helps. Better yet, adaptive-bandwidth KDE (Abramson's method) captures multi-scale spatial structure even with many observers. If a suitable saliency model exists, including it as a mixture component provides additional gains, especially with few observers. Per-image parameter optimization should be the default, particularly with variable image content or few observers. Finally, for comparability, estimation method and parameters should always be documented.
Practitioners should also choose between LOSO and LOFO cross-validation based on their benchmark design: LOSO when test subjects are genuinely different from training subjects (e.g., MIT300), LOFO when the subject pool is shared between training and test sets. In either case, properly crossvalidated estimates should replace pooled ``upper'' gold standards, which we show overfit substantially (Section~\ref{sec:experiments:upper_overfitting}).

\paragraph{Limitations.}
Including a saliency model in the empirical density may seem circular, but the concern is misplaced: the saliency model contributes cross-image information---what image features people tend to fixate on across \emph{other} images---that complements the image-specific KDE. It regularizes sparse fixation data with learned semantic priors without ever seeing the held-out fixation being evaluated.
Per-image optimization incurs computational cost, but our PyTorch implementation with GPU acceleration makes this tractable for standard datasets.
Finally, our results depend on DeepGaze MSDB quality; as saliency models improve, density estimates will improve further. This is a feature, not a bug: the gold standard should track methodological progress.

\paragraph{Conclusion.}
The ceiling for fixation prediction is substantially higher than previously estimated, restoring meaningful headroom for model improvement across images, video, and applied domains. Empirical fixation densities are evolving estimates that improve with better methodology.%

\section*{Acknowledgements}
SA was supported by the German Research Foundation (DFG): SFB 1233, Robust Vision: Inference Principles and Neural Mechanisms, TP C2, project number: 276693517.
SA thanks the International Max Planck Research School for Intelligent Systems (IMPRS-IS) for support.

\bibliographystyle{splncs04}
\bibliography{literature}

\newpage
\appendix

\section*{Appendix for ``Raising the Ceiling: Better Empirical Fixation Densities for Saliency Benchmarking''}  %

\section{Dataset Preprocessing}
\label{app:datasets}
In \cite{kummererModelingSaliencyDataset2025}, the authors reported that the CAT2000 dataset contains an artifact in the fixation data, where sometimes whole scanpaths were very concentrated in an unexpected location, which they attributed to a recording error. They therefore excluded the affected images from their analyses. We follow the same procedure and exclude the same images from our analyses on CAT2000.

\section{LOSO Density Estimation Details}
\label{app:loso}

The cross-validated model described in the main text is \emph{subject-dependent}: for each observer, a separate density is constructed from all other observers' fixations. For downstream uses requiring a single density map per image, we construct a \emph{locally crossvalidated} spatial density by combining the LOSO models via locally weighted averaging. For each subject~$s$, we define a spatial weight map~$\omega_s(\mathbf{x})$ that is~$1$ at the locations of subject~$s$'s fixations and~$0$ at all other fixation locations, with values interpolated smoothly in between. The locally crossvalidated density is then
\begin{equation}
    \label{eq:pseudo_loso}
    \hat{p}_{\mathrm{LOSO}}(\mathbf{x}) \;\propto\; \sum_{s} \bar{\omega}_s(\mathbf{x})\, p_s(\mathbf{x})\,,
\end{equation}
where $p_s(\mathbf{x})$ is the LOSO density trained on all subjects except~$s$, and $\bar{\omega}_s(\mathbf{x}) = \omega_s(\mathbf{x}) / \sum_{s'} \omega_{s'}(\mathbf{x})$ are the normalized weights. The result is renormalized to integrate to one. At each fixation location, the density is dominated by the LOSO model that \emph{excluded} that fixation's observer; away from fixation locations, the density blends across folds.

For the weight map interpolation, we explored several alternatives: linear interpolation, nearest-neighbor assignment, and radial basis function (RBF) interpolation with varying exponents. Nearest-neighbor assignment produces the correct behavior locally at fixation locations but introduces discontinuities at Voronoi cell boundaries. Linear interpolation is smooth but can produce undesirable artifacts near boundaries between groups of fixations. Squared linear RBF interpolation, more precisely $\omega_s(\mathbf{x}) \propto \max(0, 1 - \|\mathbf{x} - \mathbf{x}_s\| / r)^2$ with~$r$ set to a suitable radius---provides a smooth approximation to nearest-neighbor assignment while avoiding discontinuities, and we adopt it throughout.

\section{Density Visualization Details}
\label{app:visualization}

For the saturating heatmap, we express density as a ratio to the uniform density~$p_{\mathrm{U}} = 1/(W \cdot H)$ and apply a saturating nonlinearity
\begin{equation}
    f(d) = L\,(1 - \exp(-d/L))\,,
\end{equation}
where $d = p(\mathbf{x}) / p_{\mathrm{U}}$ is the density ratio and $L$ is a fixed saturation level. We use $L = 20$ throughout, which was experimentally choosen to provide a good dynamic range for typical fixation densities: near-uniform distributions (density ratios close to~$1$) appear pale, while more peaked distributions (density ratios of tens or more) saturate, enabling immediate visual comparison of overall peakedness across images without per-image rescaling.

For the log-spaced contour lines, we place contours at density levels $\gamma^k \cdot p_{\mathrm{U}}$ for integer~$k$, with a thicker line at~$k = 0$ (the uniform density level). We use $\gamma = 4$ in this paper, so that each contour crossing represents a factor-of-four change in density. It provides a good tradeoff between detail for high-entropy distributions (where contours are widely spaced) and low-entropy ones (where contours are closely spaced), while remaining easy to interpret.

\section{Optimization Details}
\label{app:optimization}

For each LOSO fold, the KDE is constructed from the remaining observers and the log-likelihoods of all mixture components are evaluated at the held-out fixation locations. Since the regularizer models (center bias, uniform, saliency) are precomputed, their pointwise log-likelihoods are simply looked up. The per-image objective is therefore cheap to evaluate: the KDE reduces to operations on the pairwise distance matrix between source and held-out fixations, avoiding grid-based convolutions entirely.

All parameters are optimized in log-space to ensure positivity and improve the optimization landscape. For the adaptive KDE, we additionally constrain the minimum effective bandwidth (across all per-fixation bandwidths in all LOSO folds) to stay above a threshold. Is is mainly necessary because in case of using a DeepGaze MSDB mixture component, if two subjects fixate the same or very nearby pixels, sometimes the optimization otherwise collapses the the kernel size onto individual fixation locations. This non-box constraint precludes standard L-BFGS-B; we therefore use SciPy's trust-region constrained optimizer (trust-constr). The entire computation graph---from pilot bandwidth through Abramson bandwidths to the mixture log-likelihood---is implemented in PyTorch and fully differentiable, providing exact gradients to the optimizer. This makes per-image optimization efficient and GPU-compatible.

Since per-image optimization can fall into local minima, we run each optimization 50 times from different random initial positions and retain the result with the highest cross-validated log-likelihood.

Per-image optimization allows the model to adapt the balance between image-specific fixation evidence and cross-image priors to the characteristics of each stimulus---for instance, increasing the uniform weight for images with random fixation patterns, or shrinking the KDE bandwidth for images with tightly clustered fixations.

\section{Extended Per-Image Analysis For Model Components}
\label{app:per-image-analysis}

\begin{figure}[tb]
    \centering
    \includegraphics[width=0.95\linewidth]{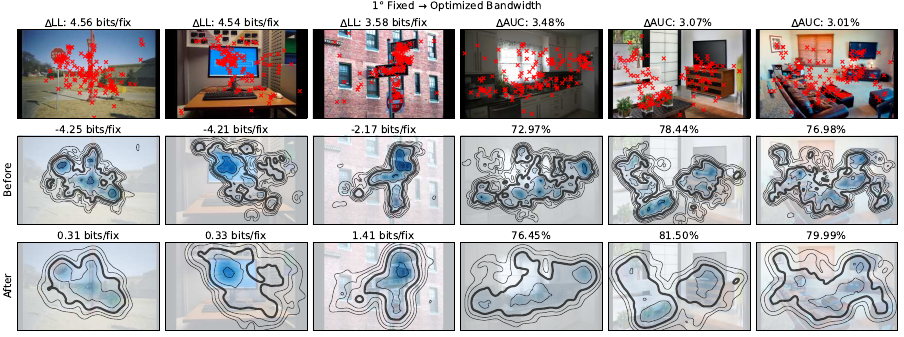}
    \caption{casestudy maximal improvement per model extension, Step 1: Starting with a classic Gaussian KDE with a bandwidth of one degree of visual angle, we change optimizing the bandwidth for leave-one-subject-out prediction and show the images where performance improves most in terms of LL and AUC.}
    \label{app:fig:casestudy_ablations_0}
\end{figure}

\begin{figure}[tb]
    \centering
    \includegraphics[width=0.95\linewidth]{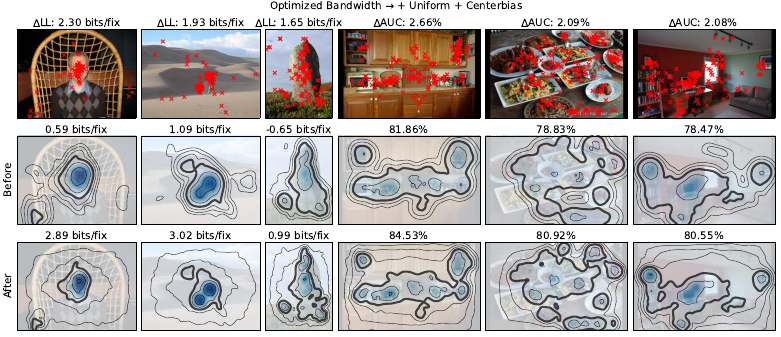}
    \caption{casestudy maximal improvement per model extension, Step 2: We now add uniform and centerbias mixture components and show the images where performance improves most in terms of LL and AUC.}
    \label{app:fig:casestudy_ablations_1}
\end{figure}

\begin{figure}[tb]
    \centering
    \includegraphics[width=0.95\linewidth]{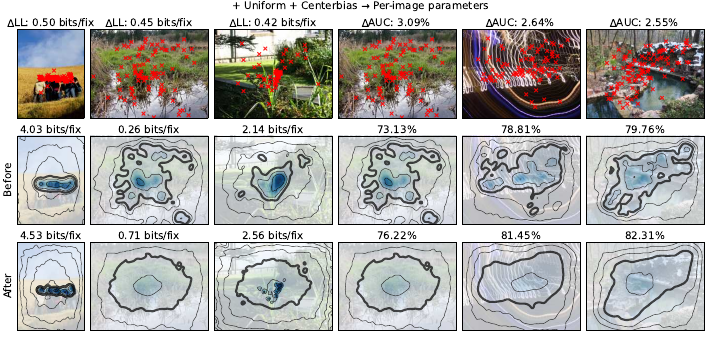}
    \caption{casestudy maximal improvement per model extension, Step 3: changing to parameters optimized per image. We show the images where performance improves most in terms of LL and AUC.}
    \label{app:fig:casestudy_ablations_2}
\end{figure}

\begin{figure}[tb]
    \centering
    \includegraphics[width=0.95\linewidth]{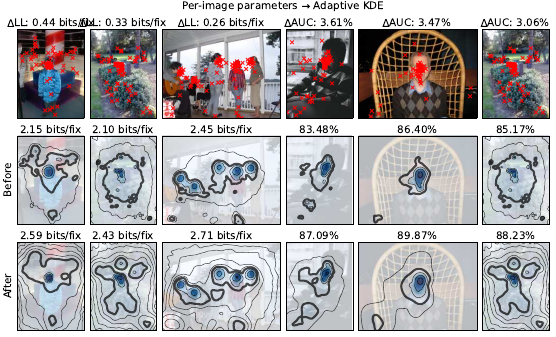}
    \caption{casestudy maximal improvement per model extension, Step 4: Switching to Abramson adaptive bandwidth KDE. We show the images where performance improves most in terms of LL and AUC.}
    \label{app:fig:casestudy_ablations_3}
\end{figure}

\begin{figure}[tb]
    \centering
    \includegraphics[width=0.95\linewidth]{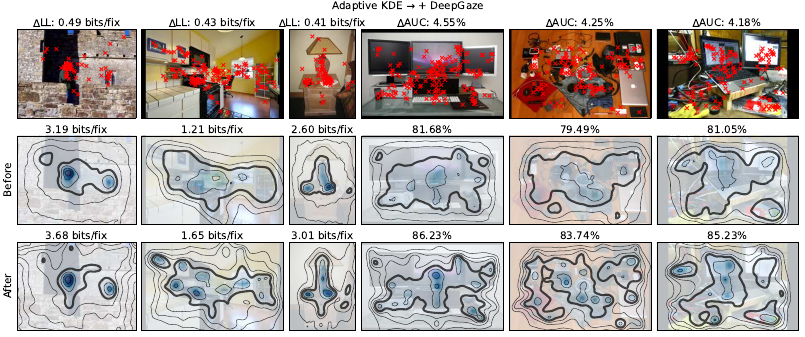}
    \caption{casestudy maximal improvement per model extension, Step 5: Adding the DeepGaze MSDB mixture component. We show the images where performance improves most in terms of LL and AUC.}
    \label{app:fig:casestudy_ablations_4}
\end{figure}

We provide extended per-image analyses for the case studies presented in the main paper. We now start wit the still most commonly used case of a Gaussian KDE with a fixed bandwidth of one degree of visual angle, and first change to a bandwidth optimized for leave-one-subject-out prediction (Figure \ref{app:fig:casestudy_ablations_0}). Next, we add the uniform and centerbias mixture components (Figure \ref{app:fig:casestudy_ablations_1}). We then change to parameters optimized per image (Figure \ref{app:fig:casestudy_ablations_2}), followed by switching to the Abramson adaptive bandwidth KDE (Figure \ref{app:fig:casestudy_ablations_3}), and finally adding the DeepGaze MSDB mixture component (Figure \ref{app:fig:casestudy_ablations_4}).

\section{Case studies by model parameter}
\label{app:case_studies_by_parameter}

Here, we go through the different model parameters (pilot bandwidth, alpha, weight of the KDE component, weight of the uniform component, weight of the centerbias component, weight of the DeepGaze MSDB component) and show for each parameter the images where this parameter has the lowest or highest value.

For the \emph{pilot bandwidth}, we see that it is largest for images with many fixations: in that case the adaptivity of the KDE is not necessary, and the KDE converges to a classic fixed-bandwidth KDE with small bandwidth (due to the small alpha). It is small for images with multiple fixation clusters, but not that many fixations

For the \emph{alpha}, which controlls the inverse proportionality between the pilot density and the effective bandwidth we see that it is smallest for images with very peaked fixation clusters, and largest for images with more dispersed fixation patterns.

For the \emph{weight of the KDE component}, we see that it is smallest (0.0) for images with no clear salient objects, where usually the centerbias and the deepgaze components takeover. On the other hand, it is largest (1.0) in cases with very strong and clear fixation clusters.

For the \emph{weight of the uniform component}, we see that it is largest (up to 0.15) for images with some very dispersed fixtions.

For the \emph{weight of the centerbias component}, we see that it is largest (up to 0.93) for images with not very clear structure in the fixation patterns, where observers indeed seem to fixate mostly central (landscape scenes and a visual search display).

Finally, for the \emph{weight of the DeepGaze MSDB component}, we see that it is lowest (0.0) for two visual search displays, i.e., artificial stimuli, and a complex image where the main object is mainly occluded --- all cases that are most likely not well predicted by DeepGaze MSDB. On the other hand, it is largest (1.0) for images with clear but complex fixation structures. In these cases, even the adaptive KDE might not be powerful enough to capture the fixation structure, since at many small but salient image areas, not enough fixations are available to result in a small bandwidth.

\begin{figure}[tb]
    \centering
    \includegraphics[width=0.95\linewidth]{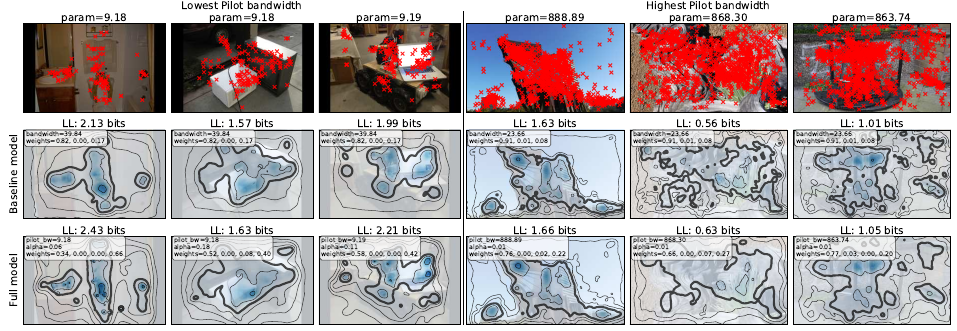}
    \caption{Extreme model parameters: Pilot bandwidth. We show the images where the pilot bandwidth is smallest or largest.}
    \label{app:fig:casestudy_extreme_parameters_pilot_bw}
\end{figure}

\begin{figure}[tb]
    \centering
    \includegraphics[width=0.95\linewidth]{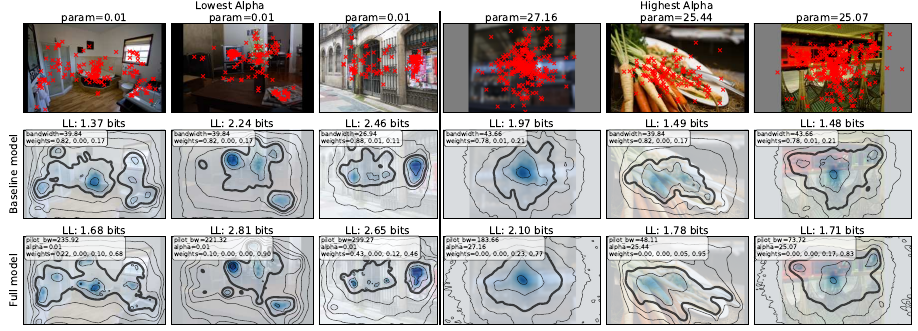}
    \caption{Extreme model parameters: Alpha. We show the images where the alpha parameter of the adaptive KDE is smallest or largest.}
    \label{app:fig:casestudy_extreme_parameters_alpha}
\end{figure}

\begin{figure}[tb]
    \centering
    \includegraphics[width=0.95\linewidth]{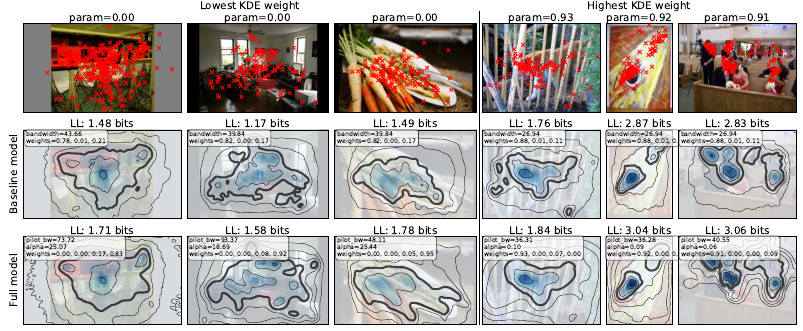}
    \caption{Extreme model parameters: Weight of the KDE component. We show the images where the weight is smallest or largest.}
    \label{app:fig:casestudy_extreme_parameters_weight_kde}
\end{figure}

\begin{figure}[tb]
    \centering
    \includegraphics[width=0.95\linewidth]{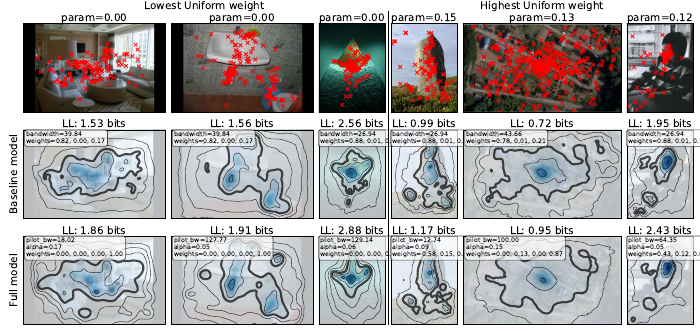}
    \caption{Extreme model parameters: Weight of the uniform component. We show the images where the weight is smallest or largest.}
    \label{app:fig:casestudy_extreme_parameters_weight_uniform}
\end{figure}

\begin{figure}[tb]
    \centering
    \includegraphics[width=0.95\linewidth]{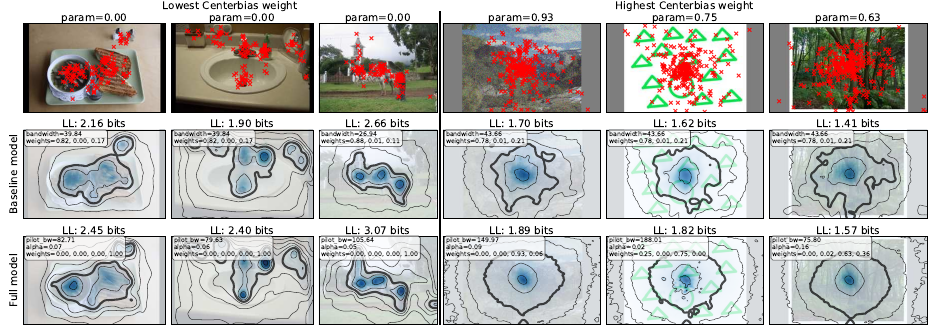}
    \caption{Extreme model parameters: Weight of the centerbias component. We show the images where the weight is smallest or largest.}
    \label{app:fig:casestudy_extreme_parameters_weight_centerbias}
\end{figure}

\begin{figure}[tb]
    \centering
    \includegraphics[width=0.95\linewidth]{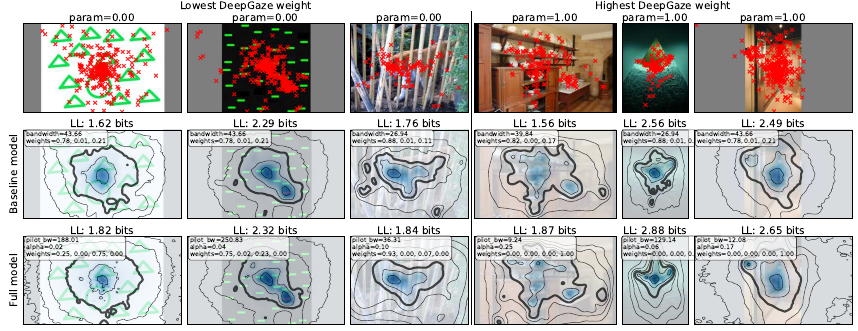}
    \caption{Extreme model parameters: Weight of the DeepGaze MSDB component. We show the images where the weight is smallest or largest.}
    \label{app:fig:casestudy_extreme_parameters_weight_deepgaze}
\end{figure}

\section{Benchmark Progress Visualization}
\label{app:benchmark_progress}

\begin{figure}[tb]
    \centering
    \includegraphics[]{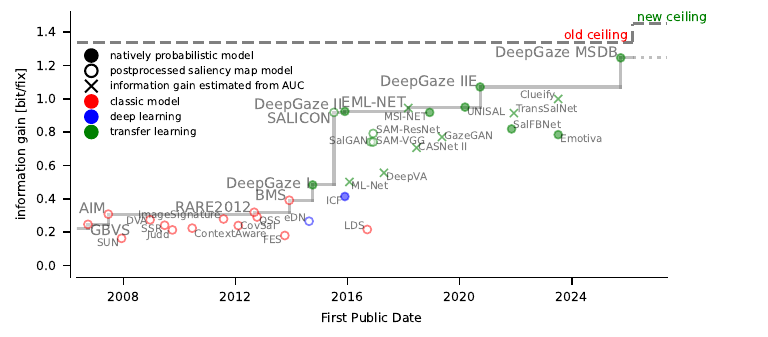}
    \caption{Progress on the MIT300 dataset of the MIT/Tuebingen Saliency Benchmark in relation to our new estimates of interobserver consistency. See \cite{kummererPredictingVisualFixations2023} for more details about model scoring.}
    \label{app:fig:benchmark_progress}
\end{figure}

In Figure~\ref{app:fig:benchmark_progress}, we visualize the progress of the MIT/Tuebingen Saliency Benchmark over time in relation to our new estimates of interobserver consistency. Applying the method of \cite{kummererInformationtheoreticModelComparison2015} and \cite{kummererPredictingVisualFixations2023}, we convert saliency map models into probabilistic models by optimizing a postprocessing consisting of blur, a pointwise monotone nonlinearity and a multiplicative centerbias for information gain on the MIT1003 dataset and apply it on MIT300. For models where no model code is available, we estimate optimal IG from AUC. See \cite{kummererPredictingVisualFixations2023} for more details.

\section{Extended Cross-Validation Design Analysis}
\label{app:cv_design}

\begin{figure}[tb]
    \centering
    \includegraphics[width=0.95\linewidth]{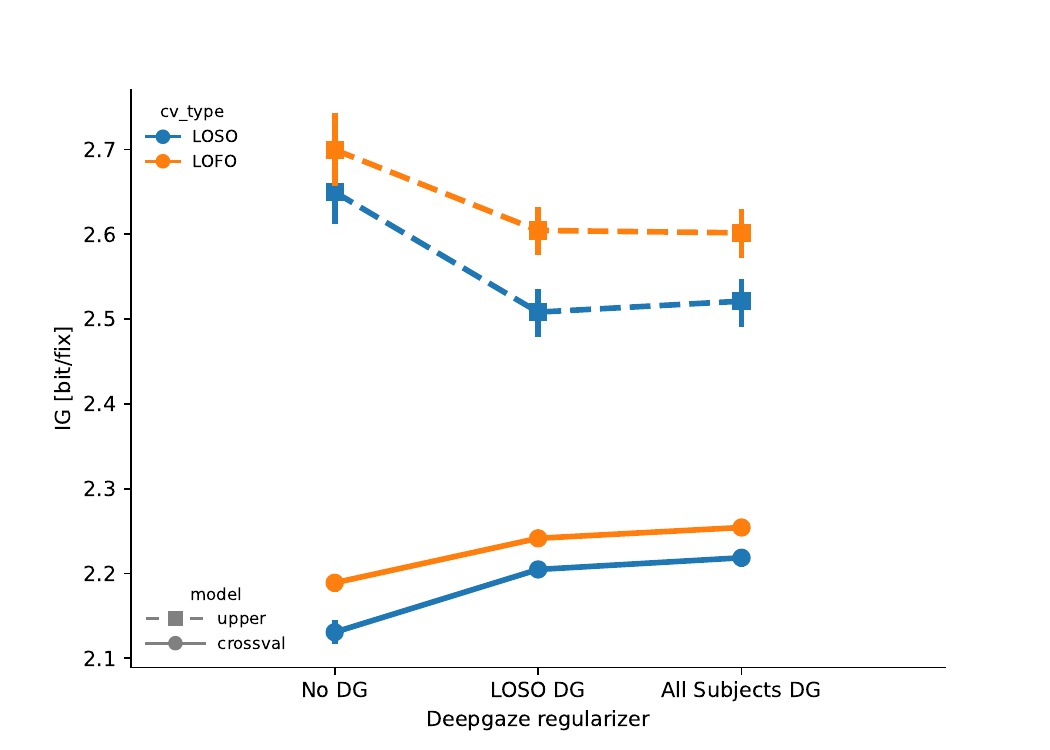}
    \caption{Extended cross-validation design analysis. Solid lines show properly crossvalidated estimates; dashed lines show pooled (``upper'') estimates. The pooled estimates overfit substantially in all configurations, with the largest overfitting without DeepGaze. The ordering between DeepGaze variants reverses between crossvalidated (solid) and pooled (dashed) evaluation, confirming that pooled estimates are dominated by self-prediction artifacts.}
    \label{app:fig:crossval_vs_upper_full}
\end{figure}

Figure~\ref{app:fig:crossval_vs_upper_full} extends the analysis of Figure~\ref{fig:cv_design}b by showing both crossvalidated and pooled (``upper'') estimates across all three DeepGaze configurations. The overfitting of pooled estimates is visible as the gap between solid and dashed lines. Two additional patterns emerge that are not visible in the main figure. First, the overfitting is most severe without a saliency model component, because the KDE receives all mixture weight. Second, the ordering between DeepGaze configurations \emph{reverses} between crossvalidated and pooled evaluation: under proper cross-validation, adding DeepGaze improves estimates (solid lines increase from left to right), while under pooled evaluation, removing DeepGaze increases scores (dashed lines decrease from left to right). This reversal confirms that pooled estimates reflect self-prediction artifacts rather than genuine predictive performance.

\section{Error Bars}

All reported error bars are bootstraped 95\% confidence intervals for the mean log-likelihood per image using the normalization method of \cite{cousineauConfidenceIntervalsWithinsubject2005} for paired comparisons with the correction of \cite{moreyConfidenceIntervalsNormalized2008}.

\end{document}